\documentclass[runningheads]{llncs}
\usepackage{graphicx}
\usepackage{amsmath}
\usepackage{amsfonts}
\usepackage{url}
\usepackage{nicematrix}
\usepackage{multirow}
\usepackage{booktabs}
\usepackage{soul}
\usepackage{color}
\usepackage{hyperref}
\begin{document}
\title{ViBERTgrid BiLSTM-CRF: Multimodal Key Information Extraction from Unstructured Financial Documents}

\titlerunning{ViBERTgrid BiLSTM-CRF}

\author{Furkan Pala\inst{1,2} \and
Mehmet Yasin Akpınar\inst{1} \and
Onur Deniz\inst{1} \and Gülşen Eryiğit\inst{2}}
\authorrunning{F. Pala et al.}
\institute{Natural Language Processing, Yapı Kredi Technology, Istanbul 34467, Turkey 
\email{\{furkan.pala,mehmetyasin.akpinar,onur.deniz\}@ykteknoloji.com.tr}\\
\url{https://ykteknoloji.com.tr/} \and
Department of Artificial Intelligence and Data Engineering, Istanbul Technical University, Istanbul 34469, Turkey\\
\email{gulsenc@itu.edu.tr}\\
\url{https://yapayzeka.itu.edu.tr/en/home}}

\maketitle              % typeset the header of the contribution
\begin{abstract}
Multimodal key information extraction (KIE) models have been studied extensively on semi-structured documents. However, their investigation on unstructured documents is an emerging research topic. The paper presents an approach to adapt a multimodal transformer (i.e., ViBERTgrid previously explored on semi-structured documents) for unstructured financial documents, by incorporating a BiLSTM-CRF layer. The proposed ViBERTgrid BiLSTM-CRF model demonstrates a significant improvement in performance (up to 2 percentage points) on named entity recognition from unstructured documents in financial domain, while maintaining its KIE performance on semi-structured documents. As an additional contribution, we publicly released token-level annotations for the SROIE dataset in order to pave the way for its use in multimodal sequence labeling models.

\end{abstract}
\section{Introduction}
\begin{figure}
\centering
\includegraphics[width=0.8\linewidth]{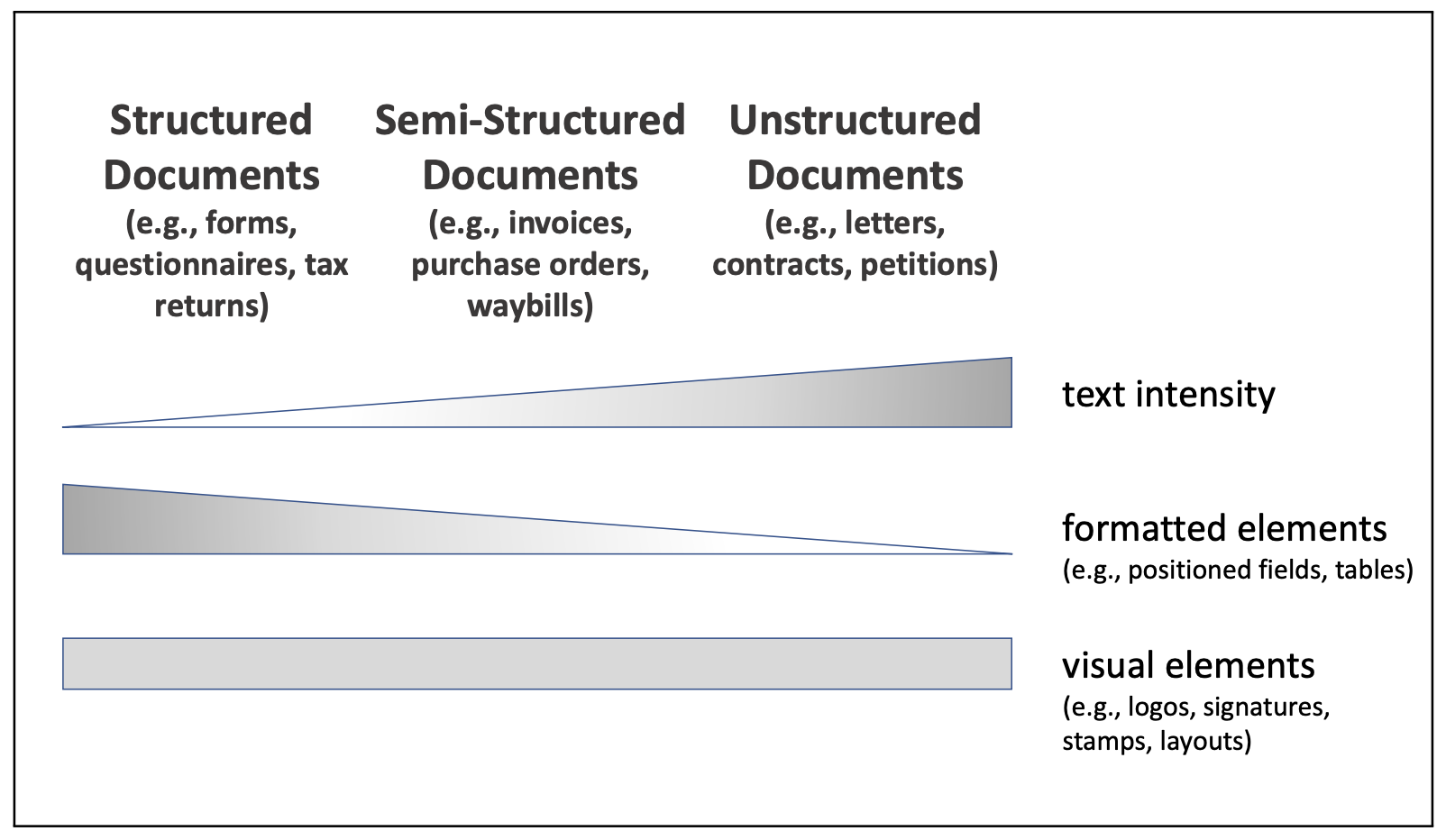}
\caption{Different types of documents~\cite{oral2022fusion}}.
\label{fig:doc_types}
\end{figure}

Documents play a crucial role in our daily lives, serving as a means of communication and record keeping. They can be written, printed, or electronic, and are often used as official records or to provide information or evidence. These documents can be categorized based on their structure and the way they present information (Fig.~\ref{fig:doc_types}).
Structured documents are highly organized, often containing tabular data and rich visual elements. Semi-structured documents have some level of organization, but may not follow a strict structure. Unstructured documents, on the other hand, do not have a predetermined structure and tend to be more text-intensive, with fewer visual or formatted elements such as tabular like layouts. 

Efficiently extracting desired information from documents through automated processing is of paramount importance, particularly in the financial domain. This need arises due to the significant amount of time and effort expended in dealing with a continuous and extensive flow of documents, a challenge that is particularly prevalent in banks. An automated process with the incorporation of both Natural Language Processing (NLP) and Computer Vision systems can significantly reduce the burden associated with document processing in the financial domain. Consequently, banks and other financial institutions can experience improved operational efficiency, enabling them to allocate resources more effectively and focus on critical tasks such as customer service and decision-making.

It is important to consider the structure and content of a document when extracting key information, as this can greatly impact the efficiency and accuracy of the extraction process. Depending on the type of document being analyzed, different approaches and architectures may be required to successfully extract the desired information. For instance, extracting information from an unstructured document may require a deep understanding of the semantics of the language, as there is no obvious predetermined structure to guide the process. This may involve the use of natural language processing techniques such as contextualized word embeddings, named entity recognition (NER) or relation extraction (RE) to identify and extract relevant information from the text. On the other hand, extracting information from a structured document may be more straightforward, as the information is typically presented in a predetermined format such as a table or form. In these cases, the extraction process may involve identifying and retrieving specific key-value pairs or cells from the document, potentially with the aid of spatial information such as the positions of fields or the layout of the document. Leveraging multimodal KIE systems can be particularly useful in this regard, as they allow us to extract information from documents using clues from textual, layout and visual features. 

In literature, there exists studies using multimodal information for document understanding. While structured and semi-structured documents have been rigorously analyzed~\cite{yu2021pick,katti2018chargrid,denk2019bertgrid,lin2021vibertgrid,kerroumi2021visualwordgrid}, extracting information from unstructured documents remains an active area of study~\cite{cristani2018future,oral2020information,oral2022fusion,stanislawer2021kleister,wyner2017deep}. This is due to the challenges of working with unstructured documents and the limited availability of datasets. Recently, with the increase in public releases of structured and semi-structured datasets~\cite{huang2019sroie,jaume2019funsd,park2019cord,riba2019table}, transformer-based multimodal models like BERTgrid and ViBERTgrid have gained popularity. These models utilize position information, which guides the process of identifying key regions in the document. However, their effectiveness has not yet been evaluated on unstructured documents.

% In literature, there exists studies which use multimodal information for document understanding. While some of these focus solely on unstructured documents~\cite{oral2020information,oral2022fusion,kleister}, some others focus on semi-structured ones~\cite{katti2018chargrid,denk2019bertgrid,lin2021vibertgrid, other Pick, layoutlm, fusiondaki citationlar...}. Since transformer based models with additional position information such as Bertgrid and ViBERTgrid became popular. However, their effectiveness has not been tested on unstructured documents yet.

In this paper, we investigate the effectiveness of multimodal transformers in understanding unstructured documents. We propose a novel architecture to enhance their performance and demonstrate its generalizability on semi-structured documents. Specifically, we introduce a new multimodal approach called ViBERTgrid BiLSTM-CRF, which combines the strengths of both transformer-based  multimodal architectures (i.e., ViBERTgrid) and sequence-based models (i.e., BiLSTM-CRF~\cite{huang2015bidirectional}). ViBERTgrid provides visual representations and learnable rich word embeddings, while BiLSTM-CRF offers syntactic and long-term context awareness. We evaluate the model on an unstructured money transfer order dataset~\cite{oral2020information} and a semi-structured receipt dataset (SROIE ~\cite{huang2019sroie}). Additionally, we introduce a new level of annotation for the SROIE dataset to be utilizable in  multimodal transformers. Our contributions may be listed as follows:
\begin{itemize}
    \item We examined the effectiveness of ViBERTgrid on unstructured documents and enhanced its performance by incorporating an additional BiLSTM-CRF layer.
    
    \item We demonstrate that the proposed architecture is generalizable to semi-structured documents.
    
    \item We publicly release token-level annotations of the SROIE dataset for use in multimodal transformers.
\end{itemize}

%As an additional contribution, we provided a new level of annotation for the SROIE dataset in order to use it in multimodal transformers (github repo for sroie dataset).

% we needed to explicitly label each token from OCR output.

% --FOOTNOTE or EXP SETUP -- Since SROIE only provides the ground truth transcriptions of key information, but does not provide the position information of labels. Thus, we pre-process the training data to generate labels of each text segment by matching their OCR results with these transcriptions.

% \paragraph{}
% -- TO BE REMOVED --
% Named entity recognition (NER) is a widely studied problem in natural language processing, which involves automatically identifying and categorizing named entities in text. These named entities can include people, organizations, locations, and other real-world objects or concepts. However, traditional NER systems often rely solely on the text of a document, ignoring other valuable sources of information such as document images and position information for OCR tokens. By incorporating these multimodal cues, we can improve the performance of NER on scanned document images, where the visual information can provide additional context and aid in the identification and classification of named entities.

%TODO: CHECK if needed
%As a result, we observed that the 
%basic ViBERTgrid was not capable of surpassing the performance of a pure textual transformer model (BERT BiLSTM-CRF) on unstructured documents.
%the introduction of our new architecture ensured that   

The paper is structured as follows:
In Section 2, we provide an overview of previous research on information extraction from documents. Section 3 delves into the specifics of the ViBERTgrid BiLSTM-CRF architecture. Section 4 details our experimental setup. Section 5 presents the results. Finally, Section 6 summarizes the key findings of our study and outlines potential avenues for future research.

\section{Related Work}
Pipelines for information extraction from documents may contain many sub-tasks including but not limited to document classification, optical character recognition (OCR), named entity recognition (NER) and relation extraction (RE). %However, we will not mention studies related to the OCR step as one can use off-the-shelf OCR engines to extract the text from document if they do not need any data specific solutions.
% NER or token based classification algorithms are crucial stages for most of the information extraction pipelines, and many studies report their performance on this task \cite{oral2020information}.
Although there exist tasks with complex n-ary nested relations \cite{oral2020information} which could only be solved by sophisticated RE algorithms~\cite{oral2019extracting,sahin2018relation,peng2017cross,jia2019document}, 
``most modern methods considered key information extraction (KIE) as a sequence tagging problem and solved through NER.''~\cite{yu2021pick}. 
However, some KIE datasets lack token-level annotations ~\cite{huang2019sroie} and researchers use in-house solutions to solve this issue \cite{yu2021pick,lin2021vibertgrid}. That is why in this study, we release a token-level annotation layer for SROIE dataset~\cite{huang2019sroie}.

Before the rise of deep neural networks, Conditional Random Fields (CRFs) were widely used for NER~\cite{lafferty2001crf,settles2004biomedical}. CRFs are statistical models that take context into account by modeling dependencies and relationships between predictions as a finite state machine. Because the NER task can be seen as a sequence labeling task, the ``linear chain'' CRFs were a popular choice since each prediction is dependent on its immediate neighbors. However, as the meaning of a word can depend not only on its immediate surroundings but also on further words, models with larger context windows have become necessary.

Deep neural networks, specifically recurrent neural networks (RNNs)~\cite{rumelhart1986rnn} and their variants, such as long short-term memory (LSTM)~\cite{Hochreiter1997lstm} and gated recurrent units (GRUs)~\cite{chung2014gru} have become the state-of-the-art models for NER. In this context, Bidirectional LSTM-CRF (BiLSTM-CRF) models~\cite{huang2015bidirectional} have been shown to be reliable and robust architectures for sequence tagging tasks as they can take advantage of both past and future tokens and capture sentence level meaning. In~\cite{oral2020information}, it was observed that replacing a simple MLP with a BiLSTM-CRF layer significantly improved the performance of BERT on the NER task. Overall, deep neural networks have been able to achieve better performance on NER tasks than pure CRF architerctures by modeling larger context windows and learning more complex relationships between predictions. 

With the introduction of large language models like ELMo~\cite{peters2018elmo}, BERT~\cite{devlin2018bert},  and GPT-3~\cite{brown2020gpt3}, the performance of NER tasks has significantly improved. These models are pre-trained in an unsupervised manner on massive amounts of text and can learn rich representations of the language. They can be fine-tuned on specific NER tasks and provide contextualized word embeddings, which capture the meaning of words in the context of a sentence. This allows the model to better understand the relationships between words and make more accurate predictions. Additionally, these models can also be used in a transfer learning setting, where the pre-trained language model is used as a feature extractor for NER tasks, further improving performance. The use of these large language models in NER tasks has become a standard approach in the field of natural language processing, achieving state-of-the-art performance on many NER benchmarks.

The usage of multimodal models using textual, layout and visual information is a promising area of research in NER task. These models are able to take advantage of the layout and visual structure of documents, such as tables and figures, to improve their understanding of the text and make more accurate predictions. For example, by incorporating information about the position of entities within a table, these models can better identify and extract entities and their relationships. This approach could be especially useful in domains such as scientific literature, where tables and figures are commonly used to present data. Further research in this area could lead to significant advancements in the field of NER.

In this regard,~\cite{katti2018chargrid}  introduced a new method called Chargrid for representing documents by converting each page into a 2D grid of characters. While this representation is useful for preserving layout and spatial features in structured documents, it may not be as effective in dealing with unstructured documents. This is because the representation relies solely on character information to encode the text, which may not be sufficient for understanding the content of the document. Nevertheless, in unstructured documents, the meaning of a character is often dependent on the surrounding context and the word it belongs to, and encoding the text using only character information can lead to a loss of important information.

Although, Chargrid representation is still useful in certain circumstances and can provide valuable insights when analyzing structured documents, it is not leveraging language models for extracting rich word embeddings. To address this deficiency,~\cite{denk2019bertgrid} modified the Chargrid architecture to represent a document as a grid of contextualized word piece embedding vectors, called BERTgrid. The BERTgrid method involves representing a document as a grid of contextualized word piece embedding vectors, which are obtained from a BERT~\cite{devlin2018bert} language model. By using this representation, the spatial structure and semantics of the document become easily accessible to the neural network processing it. This allows for more efficient and effective analysis of the document by the neural network, as it is able to easily understand the meaning and context of the words within the document. The use of contextualized embedding vectors, which contain rich information about the language and context in which words are used, further enhances the ability of the neural network to accurately process and analyze the document.

Although BERTgrid has been successful in using BERT to obtain contextualized word embeddings for key information extraction, the parameters of BERT are frozen during training, which limits the full potential of the language model. In an effort to address this issue,~\cite{lin2021vibertgrid} proposed a new architecture called ViBERTgrid and a joint-training strategy to improve the accuracy of grid-based key information extraction models. ViBERTgrid incorporates a multi-modal feature extractor, which combines BERTgrid with a CNN to process the scanned page image with the layout and textual information. Using a joint-training technique that allows for the finetuning of both the BERT and CNN models, ViBERTgrid enables BERT to learn context-specific word embeddings more effectively.

%We simplify the key information extraction process by focusing on named entity recognition, by classifying each token from the optical character recognition (OCR) output into the desired categories. However, the OCR output is often noisy, so a robust NER model is necessary. In this context, Bidirectional LSTM-CRF (BiLSTM-CRF) models~\cite{huang2015bidirectional} have been shown to be reliable and robust architectures for sequence tagging tasks. In~\cite{oral2020information}, it was observed that replacing a simple MLP with a BiLSTM-CRF layer significantly improved the performance of BERT on the NER task. On the other hand, we found that while multimodal models such as ViBERTgrid are effective for structured and semi-structured documents because they can incorporate layout and visual features, they are not as effective for unstructured documents as BERT-based BiLSTM-CRF models that use only textual features. This led us to conclude that a BiLSTM-CRF layer is essential for capturing word- and sentence-level information in long, free-form text documents. To overcome this challenge, we have designed a multimodal NER model called ViBERTgrid-BiLSTM-CRF that combines the strengths of both ViBERTgrid and BiLSTM-CRF. ViBERTgrid provides visual representations and learnable rich word embeddings, while BiLSTM-CRF offers syntactic and long-term awareness. By adding a BiLSTM-CRF layer on top of vanilla ViBERTgrid, we have created a model that is capable of effectively extracting key information from documents of all types of structures.

\section{Methodology}
The approach, as shown in Fig.~\ref{fig:vibertgrid}, consists of three main elements:

\begin{figure*}[!htp]
    \centering
    \includegraphics[width=\linewidth]{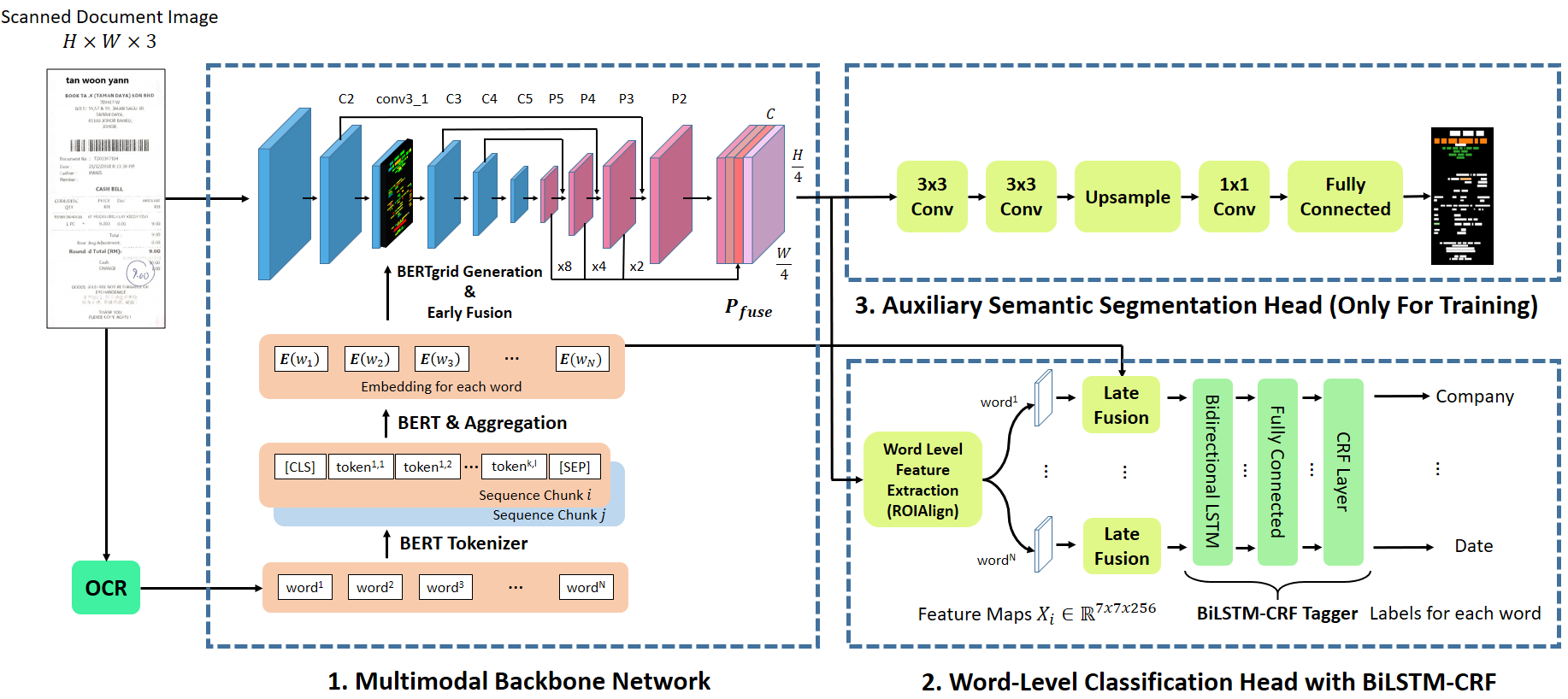}
    \caption{\textbf{ViBERTgrid BiLSTM-CRF}: Adapted architecture from ViBERTgrid.}
    \label{fig:vibertgrid}
\end{figure*}

\begin{enumerate}
    \item A multimodal backbone network that generates the ViBERTgrid feature map.
    \item A word-level classification head that predicts the label of each word. We have two variations of the word-level classification head, one with a BiLSTM-CRF layer, and the other without. In Fig.~\ref{fig:vibertgrid}, we present the one with BiLSTM-CRF.
    \item An auxiliary semantic segmentation head that performs pixel-wise classification.
\end{enumerate}
%We will provide more information on each of these modules in the following sections.

\subsection{ViBERTgrid Feature Map}
%ViBERTgrid is a rich feature map representation for documents. 
To create the ViBERTgrid representation, we first generate the BERTgrid representation and fuse it with an intermediate layer of image processing network, i.e., a backbone convolutional neural network (CNN). While BERTgrid provides textual and positional features, combining it with visual features makes the ViBERTgrid a multimodal and comprehensive embedding. %In the following subsections, we explain the BERTgrid generation and fusion steps.

\subsubsection{BERTgrid Generation}
Following~\cite{denk2019bertgrid,lin2021vibertgrid}, given a document image, OCR engine outputs words and corresponding bounding boxes. Let us represent the words as a sequence $D = \bigl[w^{(1)}, w^{(2)}, \cdots, w^{(N)}\bigr]$ of length $N$. The bounding box for each word $w^{(i)}$ is then denoted as $B^{(i)} = \Bigl(x_\textrm{min}^{(i)},y_\textrm{min}^{(i)},x_\textrm{max}^{(i)},y_\textrm{max}^{(i)}\Bigr)$ where $\Bigl(x_\textrm{min}^{(i)},y_\textrm{min}^{(i)}\Bigr)$ and $\Bigl(x_\textrm{max}^{(i)},y_\textrm{max}^{(i)}\Bigr)$ are the top-left and bottom-right corner points for the bounding box. Then, we tokenize each word in $D$ using the wordpiece (sub-word) tokenizer, resulting in a sequence of sub-word tokens denoted as $T = \bigl[t^{(1)}, t^{(2)}, \cdots, t^{(M)}\bigr]$. It is important to note that the number of sub-word tokens M is typically greater than the number of words N, as tokenization can often split a single word into multiple sub-word tokens. Because we are interested in getting word embeddings rather than token embeddings, we maintain the word indices for each token. This allows us to calculate the embedding for each word by aggregating the embeddings of its corresponding tokens. To comply with BERT's maximum input sequence length of 512, we utilize a sliding window approach to divide $T$ into chunks of 510 sub-word tokens. %This method is applied only for sequences where $M > 510$. 
Additionally, %to ensure the proper input format for BERT, 
we add [CLS] and [SEP] tokens at the beginning and end of each chunk, respectively. Lastly, we add $(510-M)$ times [PAD] tokens after the [SEP] token of the last chunk to fill it up to the required length of 512. Then, we compute the embedding $\boldsymbol{e}(t^{(i)})$ for token $t^{(i)}$ by inputting batch of chunks into the BERT language model, and average the embeddings of all tokens with the same word index to compute the  embedding $\boldsymbol{E}(w^{(j)})$ for word $w^{(j)}$. The spatial resolution of the BERTgrid is a fraction of the original document image, specifically $\frac{1}{S}$, where $S$ refers to the stride of the convolutional feature map that is fused with it. Thus, BERTgrid encoding of a document is defined as 
\[
    \boldsymbol{G}_{x,y,:} = 
    \begin{cases} 
          \boldsymbol{E}(w^{(i)}) & \textrm{if } (x \times S, y \times S) \prec B^{(i)}\\
          \boldsymbol{0}_{d} & \text{otherwise}
    \end{cases}
\]
where $\prec$ means point $(x,y)$ is inside the bounding box $B^{(i)}$. i.e., $x_\textrm{min}^{(i)} \leq x \leq x_\textrm{max}^{(i)} \land y_\textrm{min}^{(i)}\leq y \leq y_\textrm{max}^{(i)}$. $d$ is the dimension of the embedding vectors coming out of BERT. Following the above definition, we can see that the BERTgrid $\boldsymbol{G} \in \mathbb{R}^{(\frac{H}{S}) \times (\frac{W}{S}) \times d}$ is a 3-rank tensor, where $H$ and $W$ are the height and width of the document image, respectively.

\subsubsection{Early Fusion}
To incorporate BERTgrid with visual features, we merge the generated BERTgrid $\boldsymbol{G}$ with an intermediate layer of the CNN. This allows us utilize the backbone image processing CNN as a new multimodal network. As the architecture of CNN, we use a pretrained ResNet34-FPN~\cite{he2016deep} network. The early fusion process follows the same approach as in~\cite{lin2021vibertgrid} and outputs produce a highly informative multimodal feature map $\boldsymbol{P_{fuse}} \in \mathbb{R}^{(\frac{H}{4}) \times (\frac{W}{4}) \times 256}$ for document understanding tasks.

%The ResNet34-FPN outputs down-sampled feature maps $P_2, P_3, P_4, P_5$ with 256 channels and resolutions 1/4, 1/8, 1/16, 1/32 of the original document image. As mentioned earlier, we set the stride of BERTgrid tensor to 8 which allows us to concatenate it with the residual block conv3\textunderscore1 in ResNet through early fusion. Similar to~\cite{lin2021vibertgrid}, we then reduce the number of channels from 784, i.e., embedding dimension for BERT language model, to 256 through a 1x1 convolutional layer in order to apply fusion through addition with corresponding convolutional feature map. Lastly, we resize $P_3, P_4$ and $P_5$ to combine them with $P_2$ and feed through a 1x1 convolutional layer to produce a highly informative multimodal feature map $\boldsymbol{P_{fuse}} \in \mathbb{R}^{(\frac{H}{4}) \times (\frac{W}{4}) \times 256}$ for document understanding tasks.

\subsection{Word-Level Classification Head}\label{sec:word_lvl_cls}
Although key-information fields may span multiple words, the aim of this task is to classify each word into one of the predefined field labels such as \textit{company}, \textit{address}, \textit{date} and \textit{total} in the SROIE dataset. While $C$ denotes the number of predefined field labels, we also have the \textit{other} label for words we are not interested in, which makes $C+1$ labels in total. Following \cite{lin2021vibertgrid}, we use ROIAlign~\cite{he2017mask} to extract a 7x7x256 feature map $\boldsymbol{X_i}$ for each word $w^{(i)}$ from  $\boldsymbol{P_{fuse}}$ features, by aligning them with the word bounding boxes. To allow error gradients to propagate directly from the output layer to the BERT encoder layers, we pass the embedding of each word $\boldsymbol{E}(w^{(i)})$ to the word-level field type classification head through an additional skip connection. This time, we use a late fusion module in order to fuse $\boldsymbol{X_i}$ and $\boldsymbol{E}(w^{(i)})$ for each word. This module convolves each feature map $\boldsymbol{X_i}$ with two 3x3 convolutional layers, generating a new 7x7x256 feature map that is fed into a fully connected layer, which produces a 1024-d feature vector $\boldsymbol{x_i}$. We then concatenate $\boldsymbol{x_i}$ and $\boldsymbol{E}(w^{(i)})$ to obtain a $(1024+d)$-dimensional ($d$ is the dimension for word embeddings coming out of BERT) feature vector and feed this into an additional fully connected layer, which generates a new 1024-d fused feature vector $\boldsymbol{\hat{x}_i}$. To prevent overfitting, we pass $\boldsymbol{\hat{x}_i}$ through a dropout layer~\cite{srivastava2014dropout}. From this point on, we apply two different classification mechanisms: (1) using linear layers as in~\cite{lin2021vibertgrid} and (2) using a BiLSTM-CRF layer.

\subsubsection{Classification Using Linear Layers}
According to Lin et al.~\cite{lin2021vibertgrid}, incorporating two classifiers in parallel can enhance the convergence of the model. These classifiers are fully connected layers. The first classifier conducts binary classification on each word, predicting whether it belongs to any of the pre-defined $C$ labels or not. The second classifier performs multi-class classification to assign one of the $C+1$ labels to each word. It is worth noting that the first classifier only contributes to the loss during the training phase, and the second classifier is solely responsible for labeling each word during inference. To achieve this, we input a 1024-dimensional feature vector $\boldsymbol{\hat{x}_i}$ for each word $w^{(i)}$ into both classifiers in parallel, and obtain a 2-dimensional logits vector $o_1(w^{(i)})$ and a $(C+1)$-dimensional logits vector $o_2(w^{(i)})$ from the first and second classifiers, respectively. We can compute the losses for the first and second classifiers, denoted by $\mathcal{L}_{word_1}$ and $\mathcal{L}_{word_2}$, respectively, as follows:
\[
    \mathcal{L}_{word_1} = \frac{1}{N}\sum_{i=1}^{N} CELoss(o_1(w^{(i)}), y_1(w^{(i)}))
\]

\[
    \mathcal{L}_{word_2} = \frac{1}{N}\sum_{i=1}^{N} CELoss(o_2(w^{(i)}), y_2(w^{(i)}))
\]
where $y_1(w^{(i)}), y_2(w^{(i)})$ represent the ground-truth labels for the word $w^{(i)}$ for the first and second classifier, respectively. $N$ is number of words in the document detected by OCR. We used label weighting in cross entropy loss 
%Cross-entropy loss is defined as
%\[
%    CELoss(x,y) = -\sum_{c=1}^{C+1} w_c \log\frac{e^{x_c}}{\sum_{i=1}^{C+1} e^{x_c}} y_c
%\]
where $w_c$ is the weight for each label computed using the ENet weighting~\cite{paszke2016enet} as $w_c = \frac{1}{\ln(k + p_c)}$. $p_c$ is the probability of label $c$ in the train set and $k = 1.02$.
Finally, we define the word-level classification loss as $\mathcal{L}_{word} = \mathcal{L}_{word_1} + \mathcal{L}_{word_2}$.

\subsubsection{Classification Using BiLSTM-CRF Layer}
The architecture of a BiLSTM-CRF layer consists of two main components: a bidirectional LSTM (BiLSTM) layer and a conditional random field (CRF) layer. Given a list of words $D = \bigl[w^{(1)}, w^{(2)}, \cdots, w^{(N)}\bigr]$ in the document, we compute the fused feature vectors $\boldsymbol{X} = \bigl[\boldsymbol{\hat{x}_1}, \boldsymbol{\hat{x}_2}, \cdots, \boldsymbol{\hat{x}_N}\bigr]$ for each word. Then, we feed $\boldsymbol{X}$ into a BiLSTM layer to get two sequences of hidden states $\overrightarrow{\boldsymbol{H}}, \overleftarrow{\boldsymbol{H}}$
%\[
%\overrightarrow{\boldsymbol{H}} = %\bigl[\overrightarrow{\boldsymbol{h}}^{(1)}, %\overrightarrow{\boldsymbol{h}}^{(2)}, %\cdots, \overrightarrow{\boldsymbol{h}}^{(N)} % \bigr] 
%\] 
%\[
%\overleftarrow{\boldsymbol{H}} = %\bigl[\overleftarrow{\boldsymbol{h}}^{(1)}, %\overleftarrow{\boldsymbol{h}}^{(2)}, \cdots, %\overleftarrow{\boldsymbol{h}}^{(N)}  \bigr]
%\]
for forward and backward direction, respectively. We set the hidden size for the BiLSTM layer to $512$, thus the hidden state $\boldsymbol{h^{(i)}}$ for the word $w^{(i)}$ is a 512-d vector. At the end of the BiLSTM layer, these two hidden sequences are concatenated into a single sequence of hidden state vectors $\boldsymbol{H} = \bigl[\overrightarrow{\boldsymbol{H}}, \overleftarrow{\boldsymbol{H}} \bigr]$ of shape $(2 \times N, 512)$. Then, we feed $H$ into a linear layer to get emission scores matrix $\boldsymbol{E}$ of shape $(N, (C+1))$.

Following~\cite{huang2015bidirectional}, we have a CRF layer with a learnable transition matrix $\boldsymbol{T}$ of shape $((C+1),(C+1))$ where $\boldsymbol{T}_{i,j}$ represents the transition score from tag $i$ to tag $j$. Then, the score for a document $D$ is computed as 
\[
    score(D, \boldsymbol{y}) = \sum_{i=1}^{N}\boldsymbol{E}_{i, \boldsymbol{y}_i} + \sum_{i=1}^{N-1}\boldsymbol{T}_{\boldsymbol{y}_{i},\boldsymbol{y}_{i+1}}
\] 
where $y$ is an N-d vector containing the ground-truth labels. Then, the probability of predicting labels $\boldsymbol{y}$ given the document $D$ is obtained by applying normalization on all possible predictions, i.e., $\boldsymbol{\tilde{y}}$s, as 
\[
    P(\boldsymbol{y}|D)=\frac{e^{score(D, \boldsymbol{y})}}{\sum_{\boldsymbol{\tilde{y}}\in\boldsymbol{y}}e^{score(D, \boldsymbol{\tilde{y}})}}
\]
Then, the log-likelihood simplifies into 
\[
    \log P(\boldsymbol{y}|D)= score(D, \boldsymbol{y}) - \log\sum_{\boldsymbol{\tilde{y}}\in\boldsymbol{y}}e^{score(D, \boldsymbol{\tilde{y}})}
\]
During the training, the model tries to minimize the negative log-likelihood, thus our loss is $\mathcal{L}_{word} = -\log P(\boldsymbol{y}|D)$.
During the inference, we use Viterbi algorithm~\cite{viterbi} to decode the best sequence of label predictions.
\subsection{Auxiliary Semantic Segmentation Head}
As reported in~\cite{lin2021vibertgrid}, training the network with a pixel-wise semantic segmentation loss results in a faster and more stable convergence. This approach has been utilized in several architectures such as Chargrid~\cite{katti2018chargrid} and BERTgrid~\cite{denk2019bertgrid} as they rely on classifying each pixel into one of the predefined labels based on the bounding boxes of words.  Semantic segmentation head has two classifiers. The first classifies pixels into 3 categories: inside a bounding box for a word with an interested label (from $C$ categories stated in Sec.~\ref{sec:word_lvl_cls}), inside a bounding box for a word with the ``other'' label, or not in any bounding box. The second classifier directly classifies each pixel into one of $C+1$ labels. We use exactly the same structure for auxiliary semantic segmentation head 
with~\cite{lin2021vibertgrid}.

%To use these classifiers, we first arrange the dimensions of $\boldsymbol{P_{fuse}}$ by applying two 3x3 convolution layers with 256 channels and an up-sampling layer to match the resolution of feature map with the original image \hl{figurde H ve W yi belirt}, followed by two parallel 1x1 convolution layers. The first layer outputs $X_1^{out} \in \mathbb{R}^{H \times W \times 3}$ for the first classifier, while the second layer outputs $X_2^{out} \in \mathbb{R}^{H \times W \times (C+1)}$ for the second classifier. Then, we represent the output of the first and second classifiers as $o_1(x,y) \in \mathbb{R}^{3 \times 1}$ and $o_2(x,y) \in \mathbb{R}^{(C+1) \times 1}$ where $(x,y)$ represents a pixel. Therefore, the losses for the first and second classifiers, $\mathcal{L}_{seg_1}$ and $\mathcal{L}_{seg_2}$, can be calculated as:
%\small\[
%    \mathcal{L}_{seg_1} = \frac{1}{H \times W} \sum_{(x,y)\in X_1^{out}} CELoss(o_1(x,y), y_1(x,y))
%\]

%\[
%    \mathcal{L}_{seg_2} = \frac{1}{H \times W}\sum_{(x,y)\in X_2^{out}} CELoss(o_2(x,y), y_2(x,y))
%\]
%\normalsize
%where $y_1(x,y), y_2(x,y)$ denote the ground-truth categories for the pixel $(x,y)$ for the first and second classifier, respectively. $CELoss$ is the same as in the word-level classification head. Finally, we define the semantic segmentation loss as
%\[
%    \mathcal{L}_{seg} = \mathcal{L}_{seg_1} + \mathcal{L}_{seg_2}
%\]

Overall, our  introduced  architecture above differs slightly from~\cite{lin2021vibertgrid} in that 1) ResNet34-FPN was used in CNN backbone instead of ResNet18-FPN, 2) single multi-class classifier was used instead of multiple binary classifiers at word-level classification and auxiliary semantic segmentation heads, 3) ENet loss weighting strategy was used instead of positive and negative word/pixel sampling. 

%\hl{experiemnts kisminda bu degisikliklerin etkisinden bahset}.

\section{Experimental Setup}
\subsection{Datasets}
We use two types of datasets (i.e.,   semi-structured  and  unstructured) in order to evaluate the proposed approach.

\textbf{SROIE:} \label{sec:sroie} The ICDAR SROIE dataset is a publicly available collection of 973 receipts, comprising of 626 training and 347 testing samples. This dataset is highly used  in semi-structured information extraction studies and consists of four types of entities: \textit{company}, \textit{date}, \textit{address}, and \textit{total}. However, direct use of this dataset within models requiring word bounding boxes, as is the case for our models, is not trivial. %The ViBERTgrid architecture necessitates the construction of a BERTgrid representation, which requires bounding box information for each word. Unfortunately, the SROIE dataset does not provide this information. 
The ground-truth labels for key-information fields consist only of text, while the spatial information coming from OCR is separately provided for each receipt. Thus, one needs to match each word in the key-information fields with the corresponding word in OCR data. Such matching algorithms do exist~\cite{zeninglin2021vibertgridpytorch,delplace2020chargridgithub}, but they often result in poor matching, leading to suboptimal final performance. Consequently, we manually annotated the entire dataset\footnote{The annotations are available on \url{https://github.com/YKT-NLP/ICDAR-2019-SROIE-Token-Level-Annotations}} by accurately identifying and assigning key-information labels to each word using the available OCR output. This approach rendered the dataset usable for ViBERTgrid training. However, evaluating the model's performance on the test set remains problematic due to discrepancies between OCR outputs and key-information fields, such as mismatched punctuation, extra or missing white spaces between words, and typos~\cite{huang2019icdar2019}. %That's why while the model achieves quite high token-level scores, field-level scores are still poor. 
Nevertheless,~\cite{blackstar1313sroie2019} has identified and documented some of these errors over the test set, enabling us to manually rectify them and improve our model's performance. 

\textbf{Transactional Documents}: The unstructured dataset used in this study comprises two sets of Turkish money transfer order documents, previously introduced in~\cite{oral2022fusion}: UTD and UMTD. The UTD dataset contains 3500 money transfer order documents, which were selected from real banking flows. For consistency with the prior work, we used the splits provided in~\cite{oral2022fusion} as 2500 documents for training (UTD$_{train}$), 400 for validation, and 600 for testing (UTD$_{test}$). It is reported that only 7\% of UTD contains multiple transaction information within a single document. On the other hand, UMTD  contains 1154 documents which are entirely multi-transactional. We used the same splits as in the original work: 954 documents for training (UMTD$_{train}$) and 200 for testing (UMTD$_{test}$). Within the UMTD$_{test}$, \cite{oral2022fusion} identified 54 out of 200 documents with tabular-like layouts and presented the results separately for tabular-like (TLL) and non-tabular-like (noTLL) documents. Similarly, we also present our results separately to observe the effect of the BiLSTM-CRF layer on these. 
%unstructured and more structured documents.

\subsection{Hyperparameters \& Evaluation}
We pre-trained the BERT model on a vast corpus of the banking domain and utilized it in both architectures. While adopting the joint training methodology presented in~\cite{lin2021vibertgrid}, we made a modification to the optimization approach. In the original method, the CNN module was optimized using SGD while the BERT module was optimized using AdamW. However, in our implementation, we used two separate AdamW optimizers to train both the BERT and CNN models. This is because we found through our experiments that SGD often resulted in NaN gradients and was more difficult to converge compared to AdamW. The BERT optimizer has a learning rate of 5e-5 and weight decay of 0.01, while the CNN optimizer has a learning rate of 1e-4 and weight decay of 0.005. To prevent overfitting, we decayed the learning rate of both optimizers by a factor of 0.1 whenever there was no improvement in the validation micro F1 score for the past 5 epochs. We used a batch size of 2 due to large memory requirements of ViBERTgrid. As in~\cite{lin2021vibertgrid}, we used multi-scale training where the shorter size of an image is randomly selected from $\{320, 416, 512, 608, 704\}$ with a constraint that the longer side does not exceed 800. In the testing time, we set the shorter side of each image to 512 for transactional documents dataset and 704 for SROIE dataset. 

To evaluate the performance of our models, we used the official SROIE evaluation script\footnote{\url{https://rrc.cvc.uab.es/?ch=13}} to obtain the macro F1 scores on the SROIE test set. For UTD and UMTD datasets, we adopted the field-level NER F1 score evaluation methodology used by~\cite{oral2020information}. To ensure the reliability of our results, we trained and tested each model 5 times and reported the mean and standard deviation of the F1 scores. For experiments with F1 scores too far from the mean, we re-conducted them to confirm the validity of the results.

To test the significance of our results, we used McNemar's test between the merged predictions of both architectures across five runs. We selected the significance level as $\alpha=0.05$.

\section{Experiments \& Results}
In Table~\ref{tab:sroie_results}, we compare the performance of vanilla ViBERTgrid with ViBERTgrid BiLSTM-CRF architectures on the SROIE dataset. It is worth noting that we applied post-processing techniques on the predictions of both models using regular expressions from~\cite{blackstar1313sroie2019}. These techniques involve removing unwanted tokens from predictions, such as tax numbers from company names, and extracting date and total fields using rule-based approaches, such as regex pattern matching, when the model failed to predict or predictions were removed after cleaning. The post-processing techniques were applied consistently to both architectures. 

\begin{table}
\centering
\caption{Performance comparison %between vanilla ViBERTgrid and ViBERTgrid BiLSTM-CRF 
on SROIE.}\label{tab:sroie_results}
\begin{tabular}{@{}cc@{}}
\toprule
\textbf{Model}        & \textbf{Macro F1 Score (\%)} \\ \midrule
ViBERTgrid            & 93.56\tiny$\pm$0.005 \\             
ViBERTgrid BiLSTM-CRF & \textbf{93.85\tiny$\pm$0.003} \\ \bottomrule
\end{tabular}
\end{table}

\begin{table}
\centering
\caption{Field-level NER F1 Scores on two datasets: unstructured transaction document dataset (UTD) and unstructured multi-transaction document dataset (UMTD). We present the micro and macro scores in the first and second row for each cell, respectively. w/o is vanilla ViBERTgrid and w/ is ViBERTgrid BiLSTM-CRF. TLL and noTLL correspond to tabular-like and non-tabular-like documents, respectively.}\label{tab:banking_results}
\resizebox{\linewidth}{!}{%
\begin{NiceTabular}{@{}c|cc|cc|cc|cc@{}}
\toprule
\textbf{Train Set} &
\multicolumn{8}{c}{\textbf{Test Set}} \\ 
\midrule
&
\multicolumn{2}{c}{UTD$_{test}$} &
\multicolumn{6}{c}{UMTD$_{test}$} \\ 
\cmidrule{2-9} &
\multicolumn{2}{c}{All} &
\multicolumn{2}{c}{All} &
\multicolumn{2}{c}{noTLL (73\%)} &
\multicolumn{2}{c}{TLL (27\%)} \\ 
\cmidrule{2-9}
\multicolumn{1}{c}{} &
w/o &
\multicolumn{1}{c}{w/} &
w/o &
\multicolumn{1}{c}{w/} &
w/o &
\multicolumn{1}{c}{w/} &
w/o &
w/ \\ 
\midrule
UTD$_{train}$
  &
  \multicolumn{1}{c}{\begin{tabular}[c]{@{}c@{}}90.95\tiny$\pm$0.57\\ 87.76\tiny$\pm$0.53\end{tabular}} & 
  \multicolumn{1}{c}{\textbf{\begin{tabular}[c]{@{}c@{}}92.19\tiny$\pm$0.18\\ 89.57\tiny$\pm$0.39\end{tabular}}} &
  
  \begin{tabular}[c]{@{}c@{}}91.04\tiny$\pm$0.33\\ 86.71\tiny$\pm$0.82\end{tabular} &
  \multicolumn{1}{c}{\textbf{\begin{tabular}[c]{@{}c@{}}92.42\tiny$\pm$0.14\\ 89.29\tiny$\pm$1.03\end{tabular}}} &
  
  \begin{tabular}[c]{@{}c@{}}91.73\tiny$\pm$0.40\\ 85.83\tiny$\pm$1.15\end{tabular} 
  &
  \multicolumn{1}{c}{\textbf{\begin{tabular}[c]{@{}c@{}}93.03\tiny$\pm$0.19\\ 88.26\tiny$\pm$0.75\end{tabular}}}
  
  &
  \begin{tabular}[c]{@{}c@{}}89.61\tiny$\pm$0.38\\ 85.53\tiny$\pm$0.72\end{tabular}
  &
  \textbf{\begin{tabular}[c]{@{}c@{}}91.12\tiny$\pm$0.30\\ 87.97\tiny$\pm$0.70\end{tabular}}\\ \midrule
\begin{tabular}[c]%{@{}c@{}}UTD\_train (26\%) + UMTD\_train (74\%)\end{tabular} &
{@{}c@{}}UTD$_{train}$ $+$  UMTD$_{train}$ \end{tabular} 
  &
  \begin{tabular}[c]{@{}c@{}}91.05\tiny$\pm$0.30\\ 87.61\tiny$\pm$0.33\end{tabular} 
  &
  \multicolumn{1}{c}{\textbf{\begin{tabular}[c]{@{}c@{}}92.04\tiny$\pm$0.50\\ 89.09\tiny$\pm$0.53\end{tabular}}}
  
  &
  \begin{tabular}[c]{@{}c@{}}93.28\tiny$\pm$0.34\\ 90.15\tiny$\pm$0.71\end{tabular} 
  &
  \multicolumn{1}{c}{\textbf{\begin{tabular}[c]{@{}c@{}}93.98\tiny$\pm$0.29\\ 91.78\tiny$\pm$0.26\end{tabular}}}
  
  &
  \begin{tabular}[c]{@{}c@{}}93.41\tiny$\pm$0.31\\ 88.54\tiny$\pm$0.66\end{tabular} 
  &
  \multicolumn{1}{c}{\textbf{\begin{tabular}[c]{@{}c@{}}94.13\tiny$\pm$0.33\\ 89.69\tiny$\pm$0.74\end{tabular}}}
  
  &
  \begin{tabular}[c]{@{}c@{}}93.11\tiny$\pm$0.52\\ 90.66\tiny$\pm$0.65\end{tabular}
  &
  \begin{tabular}[c]{@{}c@{}}\textbf{93.70\tiny$\pm$0.70}\\ \textbf{91.77\tiny$\pm$0.57} \end{tabular}\\ \bottomrule
\end{NiceTabular}%
}

\end{table}

In  Table~\ref{tab:banking_results}, we provide a more detailed comparison on the money transaction documents. Similar to~\cite{oral2020information}, we trained vanilla ViBERTgrid and ViBERTgrid BiLSTM-CRF models on two training sets: 1) UTD$_{train}$ and 2) the union of UTD$_{train}$ and UMTD$_{train}$. 
From this point on, we refer to the experiments conducted on UTD$_{train}$ set as the first row of  Table~\ref{tab:banking_results} and
experiments conducted on UTD$_{train}$ $+$  UMTD$_{train}$ set as the second row for  simplicity. We tested each model on UTD$_{test}$ and UMTD$_{test}$. As UMTD$_{test}$ set has document splits with tabular-like (TLL) and non-tabular-like (noTLL) layouts, we give the results for these splits individually in Table~\ref{tab:banking_results}. Again, for simplicity, we refer to the UTD$_{test}$ as the first column, whole UMTD$_{test}$ as the second column, noTLL and TLL subsets of the UMTD$_{test}$ as the third and fourth columns, respectively. Thus, the table provides 8 main cells in total, each of which displays the micro and macro scores one under the other, respectively. We denoted the vanilla ViBERTgrid (without the BiLSTM-CRF layer) model as ``w/o'' and the ViBERTgrid BiLSTM-CRF as ``w/'' in this table due to space constraints.

From the results presented in Table~\ref{tab:sroie_results}, we can see that the inclusion of a BiLSTM-CRF layer resulted in a slight performance increase on the SROIE dataset. While this improvement is not statistically significant with a $p$-value of 0.24, this outcome was anticipated, as our hypothesis was that the benefits of utilizing a BiLSTM-CRF layer would be most evident on unstructured and text-heavy documents, rather than on the semi-structured receipt layouts found in the SROIE dataset.\footnote{%As explained in the Sec.~\ref{sec:sroie},Although fixing these errors helps revealing the true performance of the model, the final scores are still low compared to the literature. This is due to the fact that the state-of-the-art models use a lexicon based approach to correct the predicted results~\cite{lin2021vibertgrid,yu2021pick}.
The final scores on the SROIE dataset for both models are low compared to the literature since many of the state-of-the-art models ~\cite{lin2021vibertgrid,yu2021pick} use a lexicon-based approach to correct the predicted results.}
%as opposed to our error fixing approach introduced in Section~\ref{sec:sroie}
%Then, we utilized our in-house unstructured money transaction order document dataset to see the effectiveness of our proposed approach. 

For the unstructured transactional documents, \cite{oral2020information} reported a micro F1 score of 91.48{\tiny$\pm$0.07} and macro F1 score of 88.45{\tiny$\pm$0.09} with their pure textual model using BiLSTM-CRF on pre-trained BERT embeddings trained on UTD$_{train}$ and tested on UTD$_{test}$. However, when we evaluate the vanilla ViBERTgrid model on the same data setup (as shown in the first row and column of Table~\ref{tab:banking_results}), we observe that it underperformed, with a micro F1 score of 90.95{\tiny$\pm$0.57} and a macro F1 score of 87.76{\tiny$\pm$0.53}. We see that vanilla ViBERTgrid could not outperform a pure textual baseline. It seems that the multimodality of ViBERTgrid has a negative impact on unstructured documents, unlike its reported positive effect on semi-structured ones~\cite{lin2021vibertgrid}. We observe that the addition of a BiLSTM-CRF layer significantly improves the micro F1 score to 92.19{\tiny$\pm$0.18} and macro F1 score to 89.57{\tiny$\pm$0.39}, with a very low $p$-value of approximately 8e-15 on UTD$_{test}$.

Similarly, the effectiveness of incorporating a BiLSTM-CRF layer is also evident from the significant improvements observed in both micro and macro F1 scores on multi-transaction documents. This is clearly demonstrated by the results presented in Table~\ref{tab:banking_results}, where the second, third, and fourth columns of the first row show remarkable enhancements. In the second row of Table~\ref{tab:banking_results}, we see the results for models trained on the combination of single- and multi-transaction document collections. Likewise, BiLSTM-CRF layer significantly improves both the micro and macro F1 scores in all testing setups. %We observed significant improvements in the first three columns with $p$-values of 7e-12, 1e-5 and 0.0001, respectively. 
The fourth column, on the other hand, did not reach the significance level, as indicated by a $p$-value of 0.056, which was just above it. We interpret this as being similar to the results obtained in SROIE, where tabular-like layouts were used instead of free-format text oriented documents.

\section{Discussions \& Conclusions}
\cite{oral2022fusion} provided valuable insights into the impact of visual features on relation extraction from unstructured financial documents. However, the NER stage was left unexplored with the assumption that visual features would be more valuable at the relation extraction stage. In this paper, we focused on the impact of  using a multimodal transformer (i.e., ViBERTgrid previously explored on semi-structured documents) on the NER task from unstructured financial documents. 
However, the initial results showed that the original ViBERTgrid has a negative impact on unstructured documents compared to a pure textual baseline.
% and aimed to compare our results with previous studies conducted on unstructured banking documents (UTD and UMTD) as well as semi-structured documents (SROIE).
The paper presented an approach to enhance the performance of ViBERTgrid on unstructured documents by extending it with a BiLSTM-CRF layer. As a result, our proposed ViBERTgrid BiLSTM-CRF model demonstrated a significant improvement in performance (up to 2 percentage points) on unstructured documents, while maintaining its performance on semi-structured documents, in the domain of financial and banking documents. As an additional contribution, we publicly released token-level annotations for the SROIE dataset to pave the way for its use in multimodal sequence labeling models.

%Our study shows that multimodal transformers can also help to boost the performance of the NER stage on unstructured document understanding. However, the improvement is not as significant as the reported improvement on relation extraction with multimodal models, in comparison to pure textual transformer models.

%\section{Conclusion}

\section*{Limitations}
Although our study focused on named entity recognition task, the effectiveness of our approach on relation extraction remains unexplored. Therefore, as a future study, we plan to investigate the performance of our architecture on relation extraction and compare it with state-of-the-art multimodal relation extraction models.

\section*{Acknowledgments}
The authors want to especially thank Yapı Kredi Technology for the collection, analysis and interpretation of data from banking domain. We would like to offer our special thanks to all of our reviewers for their very valuable comments which we believe improved the final version of the article substantially. We also want to thank our colleagues Büşra Karatay and Yağmur Çağlar for their valuable discussions and support.

\section*{Ethical Statement}
In conducting this research and analyzing the in-house financial and banking data, we are committed to upholding ethical standards and ensuring the privacy and confidentiality of our customers' information. First of all, customers give open consent by sending their documents to the bank. In an everyday scenario, customers willingly send various documents such as invoices, transactions, and other relevant paperwork to the bank for the purpose of processing these documents. Staff and employees of the bank are well-trained in data privacy and protection issues. It is strictly forbidden to remove data from the secure banking network. The objective of this project is not to create recommendation systems or similar tools to alter or manipulate customer preferences. We respect the autonomy and individual choices of our customers and aim to eliminate the manual human effort of data entry as much as possible by automating the information extraction process. This will result in a significant speed-up in the services provided to customers.

\bibliographystyle{splncs04}
\bibliography{references}
\end{document}